\documentclass[11pt]{article}

\usepackage[final]{acl}

\usepackage{times}
\usepackage{latexsym}
\usepackage[T1]{fontenc}
\usepackage[utf8]{inputenc}
\usepackage{microtype}
\usepackage{inconsolata}
\usepackage{graphicx}
\usepackage{booktabs}
\usepackage{amsfonts}
\usepackage{amsmath}
\usepackage{adjustbox}
\usepackage{xcolor}
\usepackage[colorinlistoftodos]{todonotes}
\usepackage{multirow}
\usepackage{multicol}
\usepackage{tikz}
\usetikzlibrary{shapes.geometric, arrows.meta, positioning, calc}
\definecolor{badgered}{HTML}{E74C3C}
\definecolor{badgeyellow}{HTML}{E67E22}
\definecolor{badgegreen}{HTML}{27AE60}
\newcommand{\badge}[2]{%
  \tikz[baseline=(n.base)]\node[fill=#1, text=white, rounded corners=3pt,
    inner xsep=4pt, inner ysep=1.5pt,
    font=\footnotesize\sffamily\bfseries] (n) {#2};}
\newcommand{\hard}{\badge{badgered}{Hard}}
\newcommand{\med}{\badge{badgeyellow}{Med}}
\newcommand{\easy}{\badge{badgegreen}{Easy}}

\title{BenchTrace: A \textbf{Bench}mark for \textbf{T}esting \textbf{R}eflection \textbf{A}bility and \textbf{C}ontrolled \textbf{E}volution in LLM Agents}

\author{
  Jiahao Huang$^{1}$ \quad Fei Cheng$^{2,3}$ \quad Junfeng Jiang$^{3}$ \quad Zefan Yu$^{1}$ \quad Akiko Aizawa$^{1,3}$ \\[2pt]
  $^{1}$University of Tokyo \quad $^{2}$Kyoto University \quad $^{3}$National Institute of Informatics \\[2pt]
  \texttt{\{jiahao-huang, yzf930\}@g.ecc.u-tokyo.ac.jp} \quad \texttt{feicheng@i.kyoto-u.ac.jp} \\
  \texttt{\{jiang, aizawa\}@nii.ac.jp}
}

\begin{document}

\maketitle

\begin{figure*}[t]
  \centering
  \includegraphics[width=0.9\textwidth]{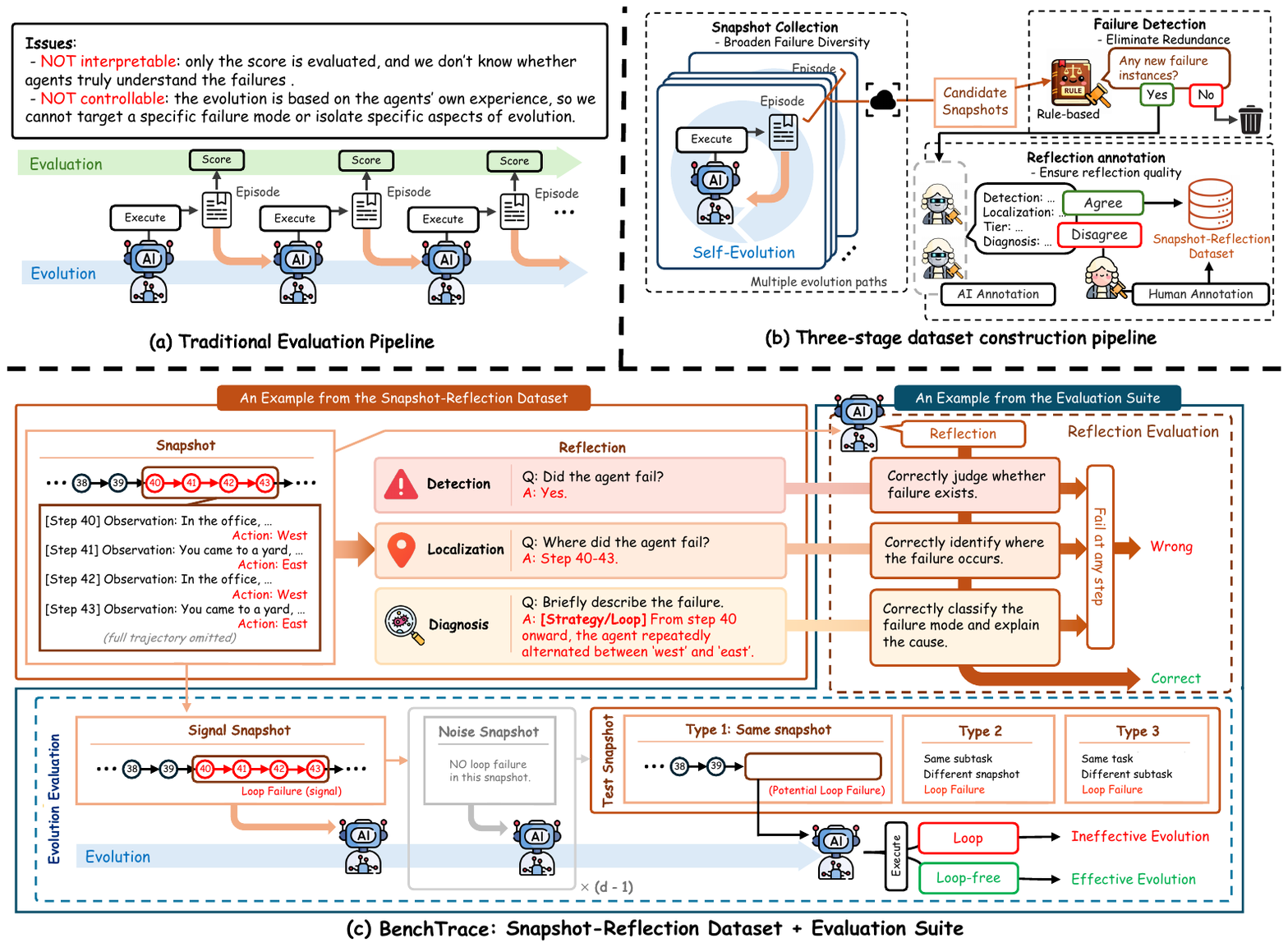}
  \caption{(a) Traditional evaluation of self-evolving agents measures only the final task score. (b) BenchTrace constructs its dataset through three stages: Snapshot Collection, Failure Detection, and Reflection Annotation. (c) BenchTrace comprises a \textbf{Snapshot-Reflection Dataset} and an \textbf{Evaluation Suite} with a \textbf{Reflection Evaluation} and an \textbf{Evolution Evaluation}.}
  \label{fig:overview}
\end{figure*}

\begin{abstract}
Self-evolving agents improve over time by reflecting on past failures, but existing evaluation is limited in two ways: it measures only task scores, leaving reflection quality unknown, and it relies on agents' own episode runs, offering no mechanism to target specific failure patterns.
We present \textbf{BenchTrace}, a benchmark for evaluating self-evolution ability in LLM agents.
BenchTrace is built on a snapshot-reflection dataset of 1,821 annotated episodes spanning six diverse tasks, and comprises a \textbf{Reflection Evaluation} that probes failure identification through targeted QA tasks, and an \textbf{Evolution Evaluation} that tests whether past failure experience translates into avoidance behavior in a controlled self-evolution simulation.
Building on BenchTrace, we propose \textbf{failure avoidance rate (FAR)}, a new evaluation metric measuring the fraction of test cases in which the agent successfully avoids the target failure instance.
Experiments with Qwen3-32B and GPT-4.1 reveal that both models fall below a 30\% end-to-end pass rate on reflection evaluation, with diagnosis as the primary bottleneck.
Evolution evaluation shows that self-evolution methods generally improve FAR over the non-evolving baseline, but agents forget early lessons as noise episodes accumulate, and agents fail to generalize their reflections beyond the specific context, causing negative transfer across task contexts.
Our correlation analysis further reveals that only a fully correct reflection is strongly associated with higher FAR.
BenchTrace exposes concrete limits of current self-evolution approaches and provides a controlled, model-agnostic framework for targeted evaluation.
\end{abstract}

\section{Introduction}
\label{sec:intro}

Large language model (LLM) agents have demonstrated remarkable capabilities across a wide range of complex, long-horizon agentic tasks~\citep{yao2022react}.
A particularly promising research direction is \emph{self-evolution}: unlike traditional LLMs that remain static after deployment, a self-evolving agent continuously improves its performance by learning from past experience and failures~\citep{shinn2023reflexion}.
Self-evolving agents improve iteratively across \emph{episodes}~\citep{gao2026survey}, where each episode is a complete, self-contained interaction trajectory between an agent and a task environment (see Appendix~\ref{app:terminology} for examples).
Self-evolving agents typically improve through non-parametric methods, such as accumulating failure lessons in external memory~\citep{shinn2023reflexion,wei2025evo,zhang2026memrl}, iteratively refining the agent's guiding prompt~\citep{he2025evotest}, and building reusable skill artifacts~\citep{yang2026autoskill}.
The self-evolution performance of an agent ultimately depends on two factors: the \textbf{base LLM}'s ability to reflect on past failures, and the \textbf{agent}'s evolution algorithm and its effectiveness.

However, existing evaluation of self-evolving agents suffers from two issues (Figure~\ref{fig:overview}(a)).
First, the \textbf{interpretability issue}: it measures only whether task scores improve, without decoupling the base LLM's reflection ability from the agent's evolution effectiveness.
Second, the \textbf{controllability issue}: since evaluation relies on the agent's own episode runs, the failure experiences it encounters cannot be controlled, making it impossible to isolate specific aspects of self-evolution.

We therefore present \textbf{BenchTrace}, a \textbf{Bench}mark for \textbf{T}esting \textbf{R}eflection \textbf{A}bility and \textbf{C}ontrolled \textbf{E}volution in LLM agents.
BenchTrace is built on a snapshot-reflection dataset.
A \emph{snapshot} is an agent-generated episode, recording the full interaction trajectory and final outcome.
A \emph{reflection} is a structured annotation of the core failure instances in the snapshot, organized along the detection, localization, and diagnosis hierarchy.
\textbf{Detection} asks whether the snapshot contains failures worth addressing.
\textbf{Localization} identifies in which steps the failure occurred.
\textbf{Diagnosis} classifies the failure mode and explains what caused it.
Figure~\ref{fig:overview}(c) shows a representative example from the dataset.

BenchTrace addresses the interpretability issue by decoupling the base LLM's reflection ability from the agent's evolution effectiveness through two independent evaluations.
The \textbf{Reflection Evaluation} directly probes whether the base LLM can identify failures in past episodes through QA tasks at each hierarchical level, isolating reflection quality from task outcomes.
The \textbf{Evolution Evaluation} goes a step further, from diagnostic to behavioral: it presents agents with controlled snapshot sequences to simulate the self-evolution process, and evaluates their performance on subsequent episodes.

BenchTrace also addresses the controllability issue, along two dimensions.
From the perspective of evolution, we can construct specific snapshot sequences to isolate particular patterns of evolution.
From the perspective of cross-model comparison, any agent framework can be evaluated on the same sequence of evolution snapshots, making the evaluation model-agnostic and enabling fair comparison across methods.

In summary, this paper makes the following contributions.\footnote{Our code is available at \url{https://github.com/Alab-NII/BenchTrace} and the dataset at \url{https://huggingface.co/datasets/huangjh16/BenchTrace}.}
\textbf{(i)} We introduce a high-quality \textbf{snapshot-reflection dataset} spanning diverse tasks, where each snapshot is annotated with the core failure instances responsible for task failure.
\textbf{(ii)} BenchTrace provides a \textbf{Reflection Evaluation} that assesses the base LLM's reflection ability through QA tasks, and an \textbf{Evolution Evaluation} that assesses the agent's evolution effectiveness through controlled snapshot sequences, together addressing the interpretability and controllability issues.
\textbf{(iii)} Applying BenchTrace to current self-evolution methods reveals two concrete limitations in current self-evolution agents: (a) agents forget early lessons as noise episodes accumulate, and (b) agents fail to generalize their reflections beyond the specific context, causing negative transfer across task contexts.
\section{Related Work}
\label{sec:related}

\subsection{Self-Evolving Agents}
\label{sec:self_evolving}

ReAct~\citep{yao2022react} augments LLMs with interleaved reasoning and action steps but accumulates no experience across episodes.
Self-evolving agents address this by enabling cross-episode learning, keeping the model frozen and evolving the agent's context or memory.
\emph{Instance-based} methods store past episodes and retrieve relevant ones at query time: RAG~\citep{lewis2020retrieval} retrieves by semantic similarity, while ReMem~\citep{wei2025evo} and MemRL~\citep{zhang2026memrl} improve upon this through meta-reasoning over stored memories and value-based retrieval policies, respectively.
\emph{Abstraction-based} methods distill past failures into verbal lessons: Reflexion~\citep{shinn2023reflexion} generates self-critiques as episodic memory, while EvoTest~\citep{he2025evotest} combines both strategies by jointly updating memory and rewriting the guiding prompt.
AutoSkill~\citep{yang2026autoskill} takes a different approach, accumulating versioned behavioral skill artifacts across sessions.

\subsection{Benchmarks for Evaluating Self-Evolving Agents}
\label{sec:benchmarks}

Existing benchmarks offer two types of tasks.
\emph{Environment-based} tasks place agents in simulated worlds requiring sequential action execution, spanning text-based games~\citep{hausknecht2020interactive}, household manipulation~\citep{shridhar2021alfworld}, grid navigation~\citep{chevalier2019babyai}, and science experiments~\citep{wang2022scienceworld}.
\emph{Information-based} tasks require handling information-intensive queries across sessions: Evo-Memory~\citep{wei2025evo} covers QA and tool-use streams, while MemoryArena~\citep{he2026memoryarena} tests cross-session retention through web shopping, travel planning, and formal reasoning.
Both types, however, rely on aggregate outcome metrics and provide no ground-truth annotations of what agents should have learned from past failures, making it impossible to assess reflection quality directly.

\section{The BenchTrace Benchmark}
\label{sec:benchmark}

\subsection{Overview}

BenchTrace consists of two components.
The \textbf{Dataset} is a collection of snapshots, each paired with ground-truth reflection annotations along the detection, localization, and diagnosis hierarchy.
The \textbf{Evaluation Suite} is built on top of the dataset and comprises two types of evaluation: a \textbf{Reflection Evaluation} that measures how well agents can understand and diagnose individual failures, and an \textbf{Evolution Evaluation} that tests whether that understanding translates into avoiding similar mistakes across carefully designed snapshot sequences.

\subsection{The Snapshot-Reflection Dataset}
\label{sec:dataset}

\subsubsection{Failure Taxonomy}
\label{sec:taxonomy}

Self-evolution is fundamentally driven by an agent's ability to reflect on its failures. We therefore begin with a precise taxonomy of failure that serves as the foundation for the rest of the paper:
\begin{itemize}
  \item \textbf{Failure class}: the top-level grouping, comprising three classes. \textbf{System} failures involve syntactic or format violations in the agent's output. \textbf{Strategy} failures reflect high-level planning deficiencies that unfold over a range of steps. The \emph{loop} failure in Figure~\ref{fig:overview}(c) is a representative case, where the agent oscillates between the same two locations without progress due to a failure to adopt an effective exploration strategy. \textbf{Operation} failures are execution errors localized to a single step. A representative case is \emph{feedback blindness}, where the agent ignores an explicit warning and proceeds with the wrong action anyway.
  \item \textbf{Failure mode}: a recurring type of mistake within a class, defined at a level of abstraction that applies across different episodes within a task. The \emph{loop} and \emph{feedback blindness} cases above are failure modes within their respective failure classes. Failure modes are task-specific, and the complete set for each task is given in Appendix~\ref{app:taxonomy}.
  \item \textbf{Failure instance}: a concrete occurrence of a failure mode at a specific location in a specific episode, identified by a step range and a description of the agent's mistake. Figure~\ref{fig:overview}(c) shows an example snapshot containing a \emph{loop} failure instance.
\end{itemize}

\subsubsection{Dataset Collection Pipeline}
\label{sec:snapshot_collection}
The dataset is designed with two objectives: to cover a broad and diverse range of failure instances, and to pair each snapshot with high-quality reflection annotations.
As shown in Figure~\ref{fig:overview}(b), we construct the dataset through a three-stage pipeline: snapshot collection, failure instance detection, and reflection annotation.

\paragraph{Snapshot collection}
For each task, we run multiple combinations of base models and self-evolving agent frameworks, each iteratively attempting the task across episodes.
When an agent's evolution stalls due to insufficient reflection ability, human annotators intervene to help pinpoint the failure and guide the agent to continue evolving.
All episodes are recorded as candidate snapshots.
This iterative human-in-the-loop approach yields a diverse collection of failure instances, including deep failures that no single agent would naturally encounter (see Table~\ref{tab:coverage} in Appendix~\ref{app:coverage}).

\paragraph{Failure detection}
We first have human annotators review a small batch of snapshots to identify common failure modes, which are then implemented as rule-based detectors.
These detectors are applied to filter all candidate snapshots, retaining only those that introduce failure instances not yet represented in the dataset.
This prevents any single failure instance from being over-represented.
For example, given the loop failure instance in Figure~\ref{fig:overview}(c), we would discard additional snapshots showing the same loop between the same locations (yard and office), seeking instead snapshots that exhibit loops in other locations or failures of a different mode.

\paragraph{Reflection annotation}
This stage aims to produce high-quality annotations for each retained snapshot.
Each snapshot is independently annotated by two AI annotators, Claude-Sonnet-4.6~\citep{anthropic2026claude} and Gemini-2.5-Flash~\citep{comanici2025gemini25pushingfrontier}, following a structured protocol described in Appendix~\ref{app:protocol}.
For each identified failure instance, annotators provide a localization, defined as a step range, and a diagnosis consisting of a failure mode and a one-sentence description of the root cause.
Annotators are also prompted to assign each failure instance a \textbf{tier}: \textbf{core} for instances that directly caused the episode to fail or prevented a significant score gain, and \textbf{marginal} for genuine errors that had limited impact on the final outcome.
Each annotator labels at most three core failure instances per snapshot, with no limit on marginal ones.
Annotators may also propose new failure modes if a failure does not fit the existing class.
For any failure instance labeled core by either annotator, a human expert adjudicates conflicts in localization, diagnosis, and tier.
Inter-annotator agreement is reported in Appendix~\ref{app:agreement}.

Together, these three stages produce the snapshot-reflection dataset. The complete entry format is described in Appendix~\ref{app:dataset_format}.

\subsection{Evaluation Suite}
\label{sec:evaluation_suite}

The evaluation suite comprises a \textbf{Reflection Evaluation} and an \textbf{Evolution Evaluation}, with representative examples shown in Figure~\ref{fig:overview}(c).
The reflection evaluation targets base models directly: given a snapshot, it probes whether the model can detect, localize, and diagnose agent failures in the collected snapshots.
The evolution evaluation targets agent frameworks: it measures whether an agent, after being exposed to specific past failure snapshots, avoids committing the same mistakes in future episodes.
This design allows BenchTrace to evaluate reflection quality and evolution improvement independently.

\subsubsection{Reflection Evaluation}

Detection, localization, and diagnosis are evaluated as separate questions, with oracle inputs provided where needed to prevent errors from cascading across levels (prompts in Appendix~\ref{app:reflection_prompts}).

\paragraph{Detection}
The model is asked whether the agent's performance has room for improvement in each snapshot.
We report \textbf{Accuracy} as the detection metric.

\paragraph{Localization}
The model is asked to identify up to three step ranges where the failure instance is located.
Ground truth consists of the step ranges of the annotated core failure instances.
Localization correctness is reported along two dimensions.
\textbf{Similarity} evaluates the accuracy of each predicted range, and \textbf{Recall} evaluates coverage of ground-truth failures.
Both are computed by matching each range to its closest counterpart via Jaccard similarity: similarity averages over predicted ranges, and recall averages over ground-truth ranges.

\paragraph{Diagnosis}
Given the ground-truth step range, the model is asked to identify the failure mode and provide a one-sentence description of the underlying failure.
Three metrics are reported: \textbf{Mode Accuracy}, whether the predicted failure mode matches the annotated one; \textbf{Token F1}, word-level F1 between the predicted and ground-truth description; and \textbf{LLM-Judge}, a score assigned by an LLM judge: 0 if the predicted description identifies a different failure, 0.5 if it partially matches the ground truth, and 1 if it correctly identifies the same underlying failure (prompt in Appendix~\ref{app:judge_prompt}).

\subsubsection{Evolution Evaluation}
\label{sec:evolution_evaluation}

In the evolution evaluation, agents evolve through \emph{evolution snapshots} and are evaluated on a \emph{test snapshot}.
While our snapshot dataset enables flexible construction of evolution snapshot sequences, we focus the evaluation on one core capability: \textbf{failure avoidance} --- whether an agent avoids making mistakes similar to those it has previously encountered.
We now describe how evolution snapshots and the test snapshot are designed toward this goal.

An evolution snapshot $s_0$ contains a failure instance $f_0$; the test snapshot $s_d$ contains a target failure instance $f_d$ of the same failure mode.
Having encountered $f_0$ in $s_0$, the agent is expected to avoid the similar failure $f_d$ in $s_d$.
\textbf{To ensure the agent learns to avoid $f_d$ solely from $f_0$}, all evolution snapshots other than $s_0$ contain no failure instances of the same failure mode as $f_d$.
We therefore call $s_0$ the \emph{signal snapshot} and the remaining ones \emph{noise snapshots}.
\textbf{To ensure the agent always faces the environment where $f_d$ can occur}, $s_d$ is truncated just before $f_d$ would occur.
Otherwise, there is no guarantee the agent ever faces the same environment.
Thus, the evolution snapshot sequence is written as $(s_0, s_1, \ldots, s_{d-1})$, and the test snapshot as $s_d^{\leq k}$, where $k$ is the last step before $f_d$ would occur.

To further analyze why agents do not avoid $f_d$, we target two typical patterns: (1) the agent forgets an earlier lesson after evolving through subsequent episodes; (2) its reflection does not generalize beyond the context of $f_0$ (see Appendix~\ref{app:failure_patterns} for an example).
We control these two patterns with \textbf{distance} and \textbf{failure proximity} respectively.
\textbf{Distance} $d$ equals the total number of evolution snapshots, of which $d-1$ are noise snapshots.
A larger $d$ tests whether the agent retains the lesson from $s_0$ under greater distraction.
\textbf{Failure proximity} controls the relationship between $s_0$ and $s_d$, with Types~1 to~3 requiring progressively stronger generalization in the agent's reflection: in Type~1 (snapshot-level), $s_0$ is the same snapshot as $s_d$; in Type~2 (subtask-level), $s_0$ is a different snapshot of the same subtask; in Type~3 (task-level), $s_0$ is from a different subtask of the same task.

As for evaluation metrics, we report \textbf{score} and \textbf{failure avoidance rate (FAR)} (see Appendix~\ref{app:avoidance_rate} for details on its computation).
Score is the conventional metric, measuring mean normalized task progress across all test snapshots; FAR measures whether the agent avoids $f_d$ in the post-truncation steps after $k$.
Notably, FAR requires pre-annotated failure instances and therefore cannot be realized through traditional benchmarks.
We validate its reliability through human and LLM annotation, achieving 81.8\% accuracy and Cohen's $\kappa = 0.538$ (see Appendix~\ref{app:avoidance_rate} for details).

\subsection{Benchmark Statistics}
\label{sec:stats}

\begin{table}[t]
\centering
\adjustbox{max width=\columnwidth}{%
\begin{tabular}{lrrrrrr}
\toprule
 & \textbf{Jer.} & \textbf{AW} & \textbf{BAI} & \textbf{SW} & \textbf{BWS} & \textbf{GTP} \\
\midrule
\multicolumn{7}{l}{\textit{Snapshot-Reflection Dataset}} \\
\midrule
Subtasks          &  6 &  6 & 20 & 10 &  5 &  4 \\
Snapshots         & 242 & 451 & 446 & 275 & 316 &  91 \\
Core/snap.        & 2.20 & 1.72 & 1.82 & 2.64 & 4.22 & 2.35 \\
Fail.\ modes      &  6 &  9 &  9 & 15 &  7 &  4 \\
Avg.\ steps       & 95.8 & 44.2 & 106.8 & 82.9 & 6.0 & 6.5 \\
Difficulty        & \hard & \med & \med & \hard & \hard & \easy \\
\midrule
\multicolumn{7}{l}{\textit{Evaluation Suite}} \\
\midrule
Refl.\ items      & 532 & 774 & 813 & 726 & 1{,}332 & 214 \\
Evol.\ T1         & 158 & 155 & 352 & 235 & 130 &  65 \\
Evol.\ T2         & 138 & 130 & 351 & 235 & 120 &  70 \\
Evol.\ T3         & 175 & 180 & 480 & 360 & 155 &  75 \\
Evol.\ total      & 471 & 465 & 1{,}183 & 830 & 405 & 210 \\
\bottomrule
\end{tabular}
}
\caption{Benchmark statistics. Jer.=Jericho, AW=AlfWorld, BAI=BabyAI, SW=ScienceWorld, BWS=Bundled Web Shopping, GTP=Group Travel Planning.}
\label{tab:benchmark_stats}
\end{table}

BenchTrace covers both task types discussed in Section~\ref{sec:benchmarks}: four environment-based tasks (Jericho~\citep{hausknecht2020interactive}, AlfWorld~\citep{shridhar2021alfworld}, BabyAI~\citep{chevalier2019babyai}, ScienceWorld~\citep{wang2022scienceworld}) and two information-based tasks (Bundled Web Shopping and Group Travel Planning~\citep{he2026memoryarena}).
Detailed statistics are provided in Table~\ref{tab:benchmark_stats}.
The snapshot-reflection dataset comprises 1,821 annotated snapshots with diverse failure profiles: failure modes range from 4 to 15, episode lengths from 6 to 107 steps, and difficulty from easy to hard (see Appendix~\ref{app:difficulty} for the difficulty criteria).
The evaluation suite totals 4,391 reflection items and 3,564 evolution test cases.

\section{Experiments}
\label{sec:experiments}

\subsection{Experimental Setup}

\paragraph{Base models}
We select one closed-source model, GPT-4.1~\citep{achiam2023gpt}, and one open-source model, Qwen3-32B~\citep{yang2025qwen3}, covering both paradigms of state-of-the-art LLMs.

\paragraph{Baseline agent frameworks}
We compare against agent frameworks from Section~\ref{sec:self_evolving} as a representative selection.
Non-evolution baselines include ReAct~\citep{yao2022react}, which does not leverage cross-episode learning.
Self-evolution baselines include RAG~\citep{lewis2020retrieval}, Reflexion~\citep{shinn2023reflexion}, EvoTest~\citep{he2025evotest}, ReMem~\citep{wei2025evo}, MemRL~\citep{zhang2026memrl}, and AutoSkill~\citep{yang2026autoskill}.

\paragraph{Prompts for Reflection Evaluation}
We illustrate the prompt design using Jericho as a representative example (Appendix~\ref{app:reflection_prompts}).
The detection question asks whether the episode contains any failure worth addressing; the localization question asks at which step range the failure occurred; the diagnosis question asks for the failure mode and a one-sentence description of its cause, given the oracle step range.
Prompts for other tasks follow the same structure and are available in our code repository.


\subsection{Main Results}
\label{sec:main_results}

\subsubsection{Reflection Evaluation Results}

\begin{table}[t]
\centering
\begin{adjustbox}{width=\columnwidth}
\begin{tabular}{lrrrrrrr}
\toprule
 & \textbf{Overall} & \textbf{Jer.} & \textbf{AW} & \textbf{BAI} & \textbf{SW} & \textbf{BWS} & \textbf{GTP} \\
\midrule
\multicolumn{8}{l}{\textit{Qwen3-32B}} \\
\midrule
Det.-Acc.  & 0.949 & 1.000 & 0.938 & 0.962 & 0.960 & 0.997 & 0.604 \\
Loc.-Sim   & 0.460 & 0.356 & 0.497 & 0.300 & 0.479 & 0.606 & 0.774 \\
Loc.-Rec   & 0.487 & 0.417 & 0.618 & 0.373 & 0.442 & 0.496 & 0.680 \\
Dia.-Mode  & 0.451 & 0.404 & 0.395 & 0.520 & 0.522 & 0.462 & 0.248 \\
Dia.-F1    & 0.339 & 0.305 & 0.322 & 0.398 & 0.359 & 0.312 & 0.262 \\
Dia.-LLM. & 0.597 & 0.669 & 0.661 & 0.709 & 0.627 & 0.522 & 0.129 \\
\midrule
\multicolumn{8}{l}{\textit{GPT-4.1}} \\
\midrule
Det.-Acc.  & 0.941 & 1.000 & 0.916 & 0.904 & 0.971 & 1.000 & 0.660 \\
Loc.-Sim   & 0.569 & 0.492 & 0.614 & 0.411 & 0.474 & 0.831 & 0.952 \\
Loc.-Rec   & 0.542 & 0.477 & 0.666 & 0.454 & 0.457 & 0.653 & 0.673 \\
Dia.-Mode  & 0.558 & 0.438 & 0.468 & 0.593 & 0.695 & 0.643 & 0.372 \\
Dia.-F1    & 0.330 & 0.299 & 0.323 & 0.386 & 0.350 & 0.329 & 0.239 \\
Dia.-LLM. & 0.754 & 0.742 & 0.793 & 0.774 & 0.755 & 0.811 & 0.289 \\
\bottomrule
\end{tabular}
\end{adjustbox}
\caption{Reflection evaluation results across six environments. LLM-Judge uses Claude Sonnet 4.6 as the judge. Acc.=Accuracy, Sim.=Similarity, Rec.=Recall.}
\label{tab:reflect_crossenv}
\end{table}

\begin{figure}[t]
\centering
\includegraphics[width=\columnwidth]{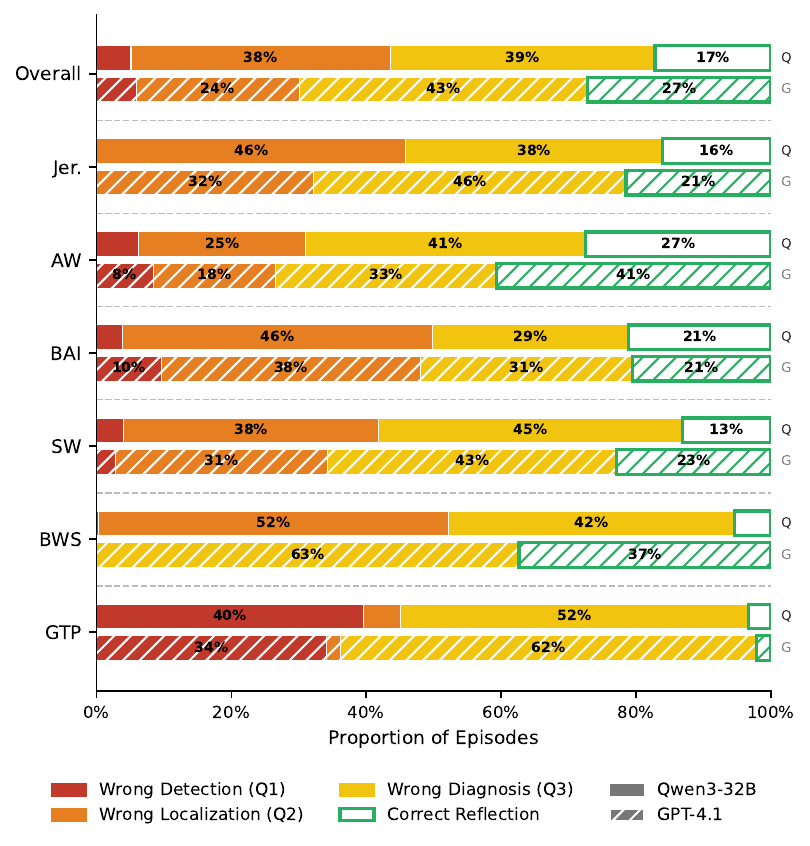}
\caption{Snapshot-level funnel analysis for reflection evaluation across six environments, showing the fraction of snapshots that pass each question in sequence. Detection is judged by binary accuracy; Localization requires Jaccard similarity $\geq 0.6$ with the ground-truth step range; Diagnosis requires the model to correctly identify the underlying failure, as judged by LLM.}
\label{fig:cascade_crossenv}
\end{figure}

We report reflection evaluation results in Table~\ref{tab:reflect_crossenv} and Figure~\ref{fig:cascade_crossenv}.
Table~\ref{tab:reflect_crossenv} reports metric scores for each question across all snapshots, with each question evaluated independently using oracle inputs from the previous level.
Figure~\ref{fig:cascade_crossenv} instead applies the three questions as an evaluation funnel, where snapshots failing detection are not evaluated on localization and those failing localization are not evaluated on diagnosis, revealing at which stage the agent's reflection breaks down.

Across both models, detection is the easiest question, with wrong detection on only 6\% of snapshots.
Both localization and diagnosis pose substantial challenges, with wrong diagnosis accounting for the largest share of wrong reflections overall.
Although wrong localization accounts for a smaller share, its consequences are more insidious: an agent that localizes to a trivial or non-failure segment will diagnose the wrong failure instance, which may actively mislead future behavior rather than improve it.
Overall, GPT-4.1 outperforms Qwen3-32B across all three questions, yet both models fall below a 30\% end-to-end pass rate, indicating that reflection ability remains a significant challenge.

Task difficulty also shapes the distribution of wrong reflections across environments.
On harder tasks such as Jericho, greater task complexity leads to more frequent and conspicuous errors, making detection relatively straightforward.
On easier tasks such as group travel planning, the snapshots we collect capture more subtle errors that models find difficult to detect, which also poses a fundamental challenge for self-evolution.

Task type further modulates the distribution of wrong reflections.
Information-based tasks such as bundled web shopping and group travel planning involve fewer steps, reducing the search space for localization and making it more tractable for stronger models.
However, the richer and more multifaceted information at each step renders precise diagnosis of the underlying failure harder, leading to more wrong diagnoses.

\subsubsection{Evolution Evaluation Results}

\begin{table*}[t]
  \centering
  \begin{adjustbox}{width=\textwidth}
  \begin{tabular}{llcccccc}
    \toprule
    \textbf{Category} & \textbf{Framework} & \textbf{Jericho} & \textbf{AlfWorld} & \textbf{BabyAI} & \textbf{ScienceWorld} & \textbf{Web Shopping} & \textbf{Travel Planning} \\
    \midrule
    Non-evolution & ReAct     & 23.5\,/\,30.1 & 50.7\,/\,15.4 &  4.9\,/\,24.0 & 26.6\,/\,51.2 &  9.2\,/\,57.8 & \textbf{41.1}\,/\,33.6 \\
    \midrule
    \multirow{6}{*}{Self-evolution}
      & RAG       & \textbf{25.0}\,/\,29.0 & 54.2\,/\,14.0 &  8.9\,/\,\textbf{40.5} & \underline{28.1}\,/\,57.6 & \underline{9.4}\,/\,\underline{60.2} & 38.0\,/\,\underline{37.1} \\
      & ReMem     & 23.8\,/\,\textbf{43.1} & \underline{55.2}\,/\,\underline{27.4} & \textbf{33.5}\,/\,14.6 & 27.9\,/\,56.2 & \underline{9.4}\,/\,58.2 & \underline{40.9}\,/\,\textbf{39.5} \\
      & MemRL     & \underline{24.5}\,/\,31.8 & \textbf{57.2}\,/\,\textbf{28.5} &  5.8\,/\,\underline{31.5} & \textbf{29.0}\,/\,\underline{60.2} &  9.3\,/\,\textbf{61.2} & 38.1\,/\,32.3 \\
      & Reflexion & 23.0\,/\,31.9 & 50.6\,/\,16.1 & \underline{13.4}\,/\,24.5 & 27.4\,/\,57.9 &  8.6\,/\,53.7 & 38.0\,/\,35.9 \\
      & EvoTest   & 23.2\,/\,\underline{38.1} & 52.4\,/\,21.3 & 10.3\,/\,27.6 & \underline{28.1}\,/\,\textbf{64.5} & \textbf{10.5}\,/\,58.8 & 36.5\,/\,37.1 \\
      & AutoSkill & 23.6\,/\,30.8 & 53.0\,/\,17.1 &  7.1\,/\,30.0 & 26.2\,/\,51.0 &  7.3\,/\,53.8 & 38.5\,/\,35.9 \\
    \bottomrule
  \end{tabular}
  \end{adjustbox}
  \caption{Evolution evaluation results with Qwen3-32B across agent frameworks and six tasks. Each cell reports score\,/\,FAR (\%). Score is progress rate for all tasks except BabyAI, which reports win rate. \textbf{Bold} and \underline{underline} denote the best and second-best per column.}
  \label{tab:main_results}
\end{table*}

Table~\ref{tab:main_results} shows the evolution evaluation results with Qwen3-32B across six tasks.
In most cases, self-evolution methods improve over the non-evolution baseline ReAct on both score and FAR, indicating that effective reflection enables agents to achieve higher task performance and avoid previously encountered failure instances.
An exception is group travel planning, where ReAct achieves the highest score among all methods.
We attribute this to the poor reflection quality on this task: as shown in Figure~\ref{fig:cascade_crossenv}, group travel planning has the lowest correct reflection rate across all environments, and self-evolution can even have a negative effect on performance.
A deeper analysis of the correlation between reflection quality and evolution performance is provided in Section~\ref{sec:reflection_avoidance_corr}.

Score variance across methods is relatively small, especially on harder tasks such as Jericho and ScienceWorld.
\textbf{FAR captures evolution signals that score alone cannot.}
This is especially pronounced on harder tasks: on Jericho, score spans only 2 points across all methods while FAR spans 13.
On harder tasks, achieving a score improvement typically requires a long sequence of correct operations, making it inherently unlikely; yet if an agent successfully avoids previously encountered failures, we argue this constitutes meaningful self-evolution even when the score remains flat.

We select Jericho and AlfWorld as representative high- and medium-difficulty tasks and conduct additional experiments with GPT-4.1, reported in Table~\ref{tab:gpt41_results}.
GPT-4.1 achieves higher overall performance than Qwen3-32B, benefiting from its stronger reflection and execution capabilities.
However, we find that the FAR of GPT-4.1 is not substantially higher than that of Qwen3-32B. We therefore report the failure repeat count (FRC $\downarrow$), the number of times $f_d$ reappears in the post-truncation steps, in Table~\ref{tab:repeat_failure} as a supplementary reference.
Under most frameworks, GPT-4.1 exhibits a lower FRC than Qwen3-32B, suggesting that GPT-4.1 is better at combining both cross-episode and in-episode failure experiences to avoid repeated mistakes.

\begin{table}[t]
  \centering
  \begin{adjustbox}{width=\columnwidth}
  \begin{tabular}{llcc}
    \toprule
    \textbf{Category} & \textbf{Framework} & \textbf{Jericho} & \textbf{AlfWorld} \\
    \midrule
    Non-evolution & ReAct     & 25.5\,/\,31.3 & 49.3\,/\,18.5 \\
    \midrule
    \multirow{6}{*}{Self-evolution}
      & RAG       & 27.9\,/\,22.2 & 50.6\,/\,14.8 \\
      & ReMem     & \underline{28.2}\,/\,30.7 & \underline{64.6}\,/\,\underline{29.2} \\
      & MemRL     & \textbf{31.0}\,/\,29.6 & \textbf{64.8}\,/\,\textbf{35.8} \\
      & Reflexion & 27.7\,/\,30.0 & 52.7\,/\,16.5 \\
      & EvoTest   & 26.4\,/\,\underline{31.7} & 47.1\,/\,17.2 \\
      & AutoSkill & 25.8\,/\,\textbf{32.2} & 52.2\,/\,16.8 \\
    \bottomrule
  \end{tabular}
  \end{adjustbox}
  \caption{Evolution evaluation results with GPT-4.1.}
  \label{tab:gpt41_results}
\end{table}

\subsection{Analysis}
\label{sec:analysis}

\subsubsection{Correlation between Reflection and Failure Avoidance}
\label{sec:reflection_avoidance_corr}

We examine whether a correct reflection on $s_0$ translates into failure avoidance on the test snapshot.
Following the Detection$\to$Localization$\to$Diagnosis funnel, Table~\ref{tab:reflection_avoidance} reports FAR for operation failures. Each row defines a cumulative correct group: Det.\ = detection correct; Loc.\ = detection and localization both correct; Diag.\ = all three correct. FAR\textbar correct and FAR\textbar wrong report the FAR of the correct group and the rest, respectively.

\begin{table}[h]
  \centering
  \begin{adjustbox}{width=\columnwidth}
  \begin{tabular}{lrrrrr}
    \toprule
    \textbf{Level} & \textbf{$n$} & \textbf{FAR\textbar correct} & \textbf{FAR\textbar wrong} & \textbf{$p$} & \textbf{Sig.} \\
    \midrule
    Det.  & 3302 & 0.385 & 0.250 & 0.4324      & \\
    Loc.  &  745 & 0.346 & 0.396 & 0.0139      & $^{*}$ \\
    Diag. &  363 & 0.485 & 0.373 & ${<}0.0001$ & $^{***}$ \\
    \bottomrule
  \end{tabular}
  \end{adjustbox}
  \caption{FAR for operation failures across six tasks, split by cumulative reflection correctness. $n$ = size of the correct group.}
  \label{tab:reflection_avoidance}
\end{table}

These results show that \textbf{a fully correct reflection, i.e., all three of detection, localization, and diagnosis correct, significantly improves FAR} ($p < 0.0001$).
Answering only detection or only localization correctly is insufficient: detection is near-universal and shows no predictive value, while correct localization alone is negatively associated with avoidance, as pinpointing the failure step without a correct diagnosis provides no actionable guidance and may even be misleading.

\subsubsection{Effect of Failure Proximity and Distance}

The results in Section~\ref{sec:main_results} report aggregate performance across all types of failure proximity and distances.
We now turn to a finer-grained analysis along the two dimensions defined in Section~\ref{sec:evolution_evaluation}: failure proximity and evolution distance $d$.
Table~\ref{tab:type_distance} breaks down FAR by failure proximity and distance on Jericho with Qwen3-32B.

\begin{figure}[t]
  \centering
  \includegraphics[width=\columnwidth]{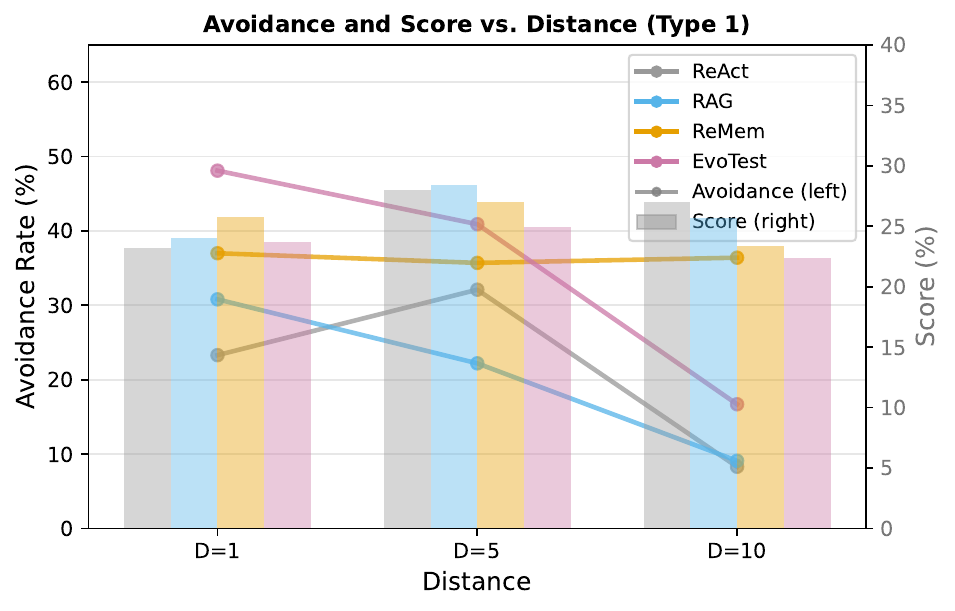}
 {\small (a) Effect of distance (Type~1)}\\[4pt]
  \includegraphics[width=\columnwidth]{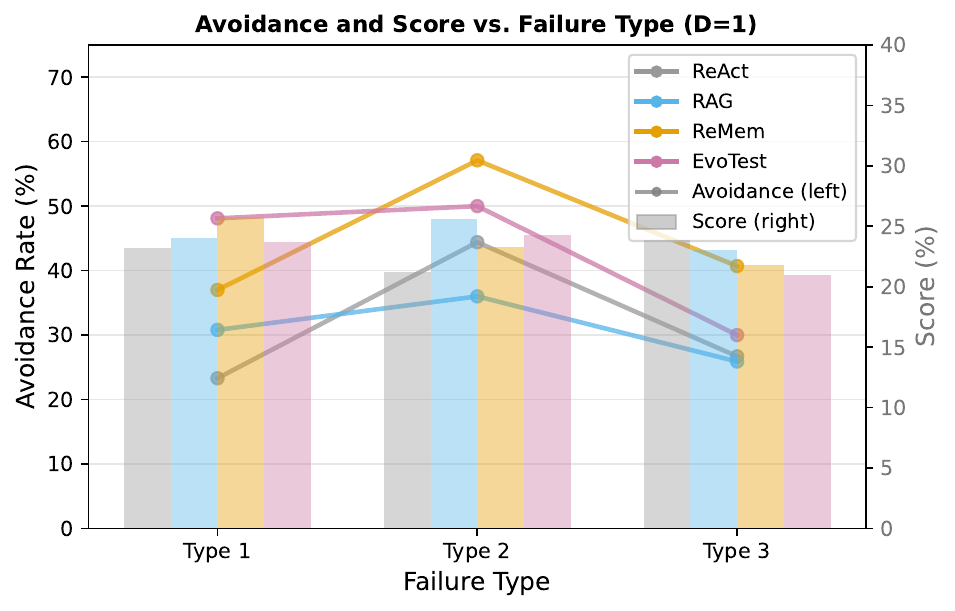}
  {\small (b) Effect of failure proximity ($d=1$)}
  \caption{FAR (line) and score (bar) on Jericho with Qwen3-32B.}
  \label{fig:analysis}
\end{figure}

As shown in Figure~\ref{fig:analysis}(a), across most methods, the FAR at $d=10$ is substantially lower than at $d=1$, confirming that \textbf{agents do forget early lessons as the number of intervening noise episodes increases}.
At short to medium distances ($d=1$ and $d=5$), EvoTest, which combines naive retrieval with prompt revision, achieves the highest FAR.
At long distance ($d=10$), however, naive retrieval is insufficient: the large number of noise episodes dilutes the memory pool, making it difficult to surface the relevant lesson.
ReMem, by selectively consolidating episodic memories, achieves the highest FAR at $d=10$ on Type~1 (36.4\%), demonstrating that principled memory management is key to long-distance self-evolution.

Beyond distance, \textbf{overly specific reflections fail to generalize across task contexts}, as shown in Figure~\ref{fig:analysis}(b): score declines monotonically from Type~1 to Type~3.
At short distance ($d=1$) under Type~3, all evolution methods score below the ReAct baseline (23.9), despite achieving positive FAR.
This indicates a form of negative transfer: the agent remembers the specific failure instance yet applies task-specific experience too rigidly, causing past lessons to become a burden rather than an asset.
As distance increases and the agent encounters more diverse episodes, this negative transfer effect weakens and the score gap between evolved methods and ReAct narrows.

\section{Conclusion}
\label{sec:conclusion}

We presented BenchTrace, a benchmark for evaluating self-evolution ability in LLM agents, built on a snapshot-reflection dataset of 1,821 annotated episodes spanning six diverse tasks.
BenchTrace comprises a Reflection Evaluation that probes failure identification through QA tasks, and an Evolution Evaluation that tests whether past failure experience translates into avoidance behavior in a controlled self-evolution simulation.

Our experiments reveal that current LLMs fall below a 30\% end-to-end pass rate on reflection evaluation, with diagnosis as the primary bottleneck, indicating that reflection ability remains a significant open challenge.
On evolution evaluation, self-evolution methods generally improve over the non-evolving baseline, yet our fine-grained analysis exposes two concrete failure modes: agents forget early lessons as the number of noise episodes increases, and overly specific reflections cause negative transfer when experience must generalize across task contexts.
Our correlation analysis further establishes a strong link between reflection quality and FAR, showing that only a fully correct reflection significantly improves avoidance.
These findings highlight the gap between current self-evolution methods and robust learning from failure, and point toward principled memory management and generalization-aware reflection as key directions for future work.

\section*{Limitations}

While BenchTrace represents a step toward more controllable and interpretable evaluation of self-evolving agents, several limitations remain.

BenchTrace currently focuses on non-parametric self-evolution methods.
Parametric approaches, such as supervised fine-tuning~\citep{yuan2025agent,ge2025samule,wang2025steca} and reinforcement learning~\citep{peng2026sage}, represent a complementary paradigm but operate at a fundamentally different scale: they typically require thousands of training episodes before yielding improvements, whereas non-parametric methods tend to saturate within tens of episodes.
This scale mismatch makes direct comparison difficult.
We explored using BenchTrace to provide a simulated evolution process for parametric methods, but the limited number of available snapshots led to substantially degraded performance.
We therefore plan to prioritize efficient and high-quality snapshot-reflection pair collection in future work, with the goal of scaling BenchTrace to support evaluation of parametric methods as well.

BenchTrace could be further scaled in both benchmark coverage and experimental breadth.
On the benchmark side, the current focus on structured task environments leaves room for extension to more open-ended real-world settings such as software engineering or web browsing, which may exhibit qualitatively different failure patterns; we leave this for future work.
On the experimental side, due to funding constraints, GPT-4.1 experiments are limited to Jericho and AlfWorld rather than the full six-task suite.

The snapshot-reflection dataset is annotated through a collaboration between AI annotators and human experts, which improves efficiency but may introduce inaccuracies.
Human adjudication is applied to all core failure instances to resolve conflicts in localization, diagnosis, and tier, providing a degree of quality control.
However, marginal failures and finer-grained annotation details remain AI-generated and are more susceptible to error.

\bibliography{references}


\appendix

\section{Terminology}
\label{app:terminology}

\begin{figure}[h]
  \centering
  \includegraphics[width=\columnwidth]{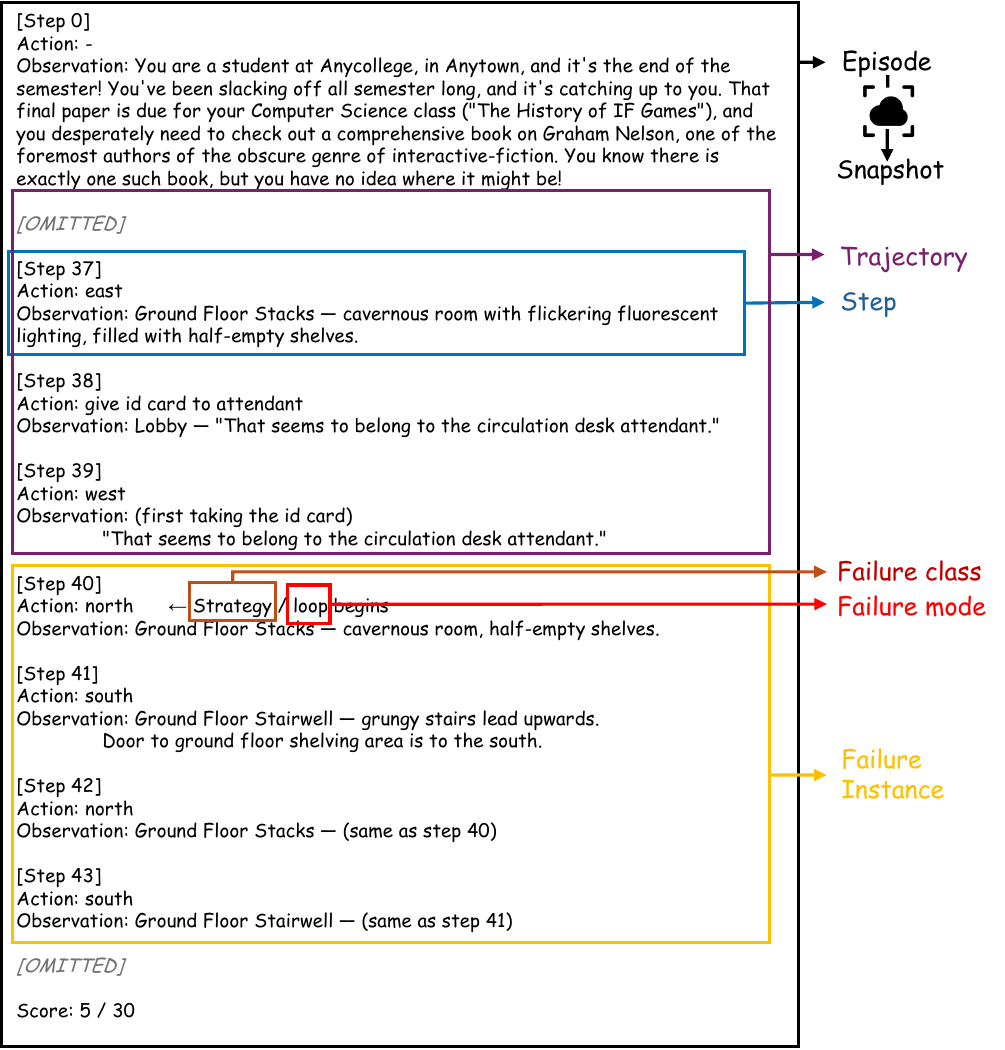}
  \caption{Key terminology illustrated with a Jericho example.}
  \label{fig:terminology_example}
\end{figure}

\begin{figure}[h]
  \centering
  \includegraphics[width=\columnwidth]{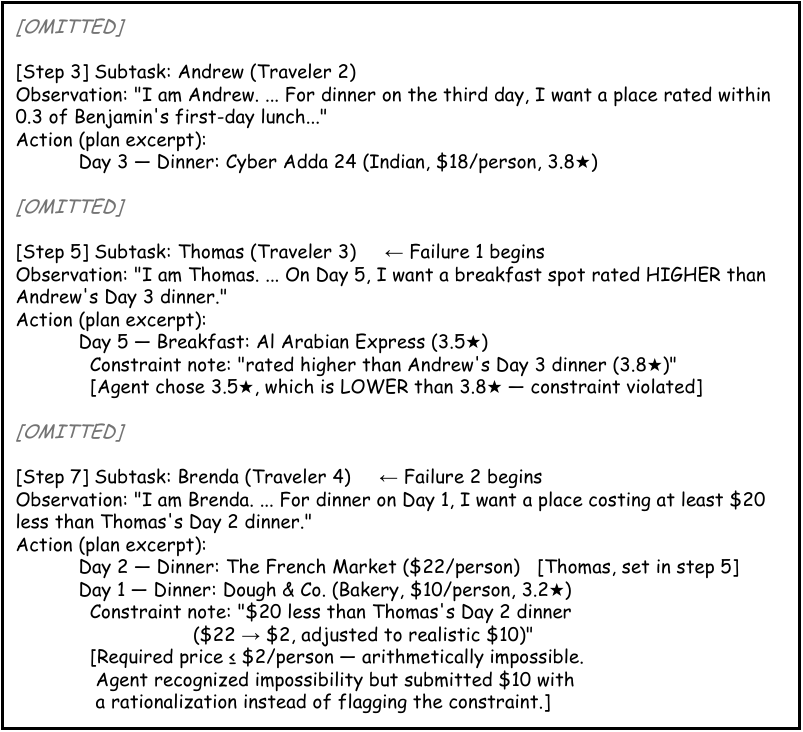}
  \caption{Key terminology illustrated with a GroupTravelPlanning example.}
  \label{fig:terminology_example_gtp}
\end{figure}

We first define the key terms used throughout this paper.
Figures~\ref{fig:terminology_example} and~\ref{fig:terminology_example_gtp} show concrete examples from Jericho and GroupTravelPlanning, representative of the environment-based and information-based task categories in BenchTrace, respectively.

\begin{itemize}
  \item \textbf{Episode.} A complete, self-contained interaction between an agent and a task environment, from the initial observation to termination. Each episode produces a sequence of steps and a final outcome.

  \item \textbf{Snapshot.} A stored, static record of an episode, containing its full trajectory and final outcome. The term \emph{episode} refers to the interaction as it unfolds dynamically; \emph{snapshot} refers to the same object once it has been recorded and can be replayed or inspected.

  \item \textbf{Step.} A single action--observation pair within an episode: the agent issues an action, and the environment returns an observation in response.

  \item \textbf{Trajectory.} The ordered sequence of steps constituting an episode, recording the full history of agent--environment interaction.

  \item \textbf{Failure class.} The coarse-grained category to which a failure belongs. BenchTrace defines three failure classes: system failures (caused by environment or interface constraints), strategy failures (flawed high-level planning or decision-making), and operation failures (incorrect low-level actions).

  \item \textbf{Failure mode.} A recurring pattern of failure within a failure class, defined as a named, reusable description of how agents typically go wrong in a particular way (e.g., action-space confusion, goal repetition, or navigation loop).

  \item \textbf{Failure instance.} A concrete occurrence of a failure mode in a specific snapshot, anchored to a contiguous range of steps where the failure manifests.
\end{itemize}

\section{Benchmark Creation}
\label{app:creation}

\subsection{Failure Instance Coverage}
\label{app:coverage}

Table~\ref{tab:coverage} compares the number of failure instances a single agent naturally encounters against the full BenchTrace dataset, which spans multiple models, algorithms, and human-guided collection rounds.

\begin{table}[h]
\centering
\small
\begin{adjustbox}{max width=\columnwidth}
\begin{tabular}{lrrrrrr}
\toprule
 & Jer. & AW & BAI & SW & BWS & GTP \\
\midrule
Single Agent & 886 & 239 & 964 & 1044 & 965 & 206 \\
\midrule
\multirow{2}{*}{BenchTrace} & 1785 & 1011 & 2316 & 1603 & 1708 & 232 \\
 & {\footnotesize\textcolor{badgegreen}{(+101\%)}} & {\footnotesize\textcolor{badgegreen}{(+323\%)}} & {\footnotesize\textcolor{badgegreen}{(+140\%)}} & {\footnotesize\textcolor{badgegreen}{(+54\%)}} & {\footnotesize\textcolor{badgegreen}{(+77\%)}} & {\footnotesize\textcolor{badgegreen}{(+13\%)}} \\
\bottomrule
\end{tabular}
\end{adjustbox}
\caption{Failure instances naturally encountered by a single agent (Qwen3-32B with EvoTest) versus the full BenchTrace dataset. Jer.=Jericho, AW=AlfWorld, BAI=BabyAI, SW=ScienceWorld, BWS=Bundled Web Shopping, GTP=Group Travel Planning.}
\label{tab:coverage}
\end{table}

The substantial increase in coverage reflects the greater diversity achieved by combining multiple models, agent frameworks, and human-guided annotation rounds, ensuring that BenchTrace captures a wider range of failure patterns than any single agent would naturally encounter.

\subsection{Annotation Guideline Prompt}
\label{app:protocol}

Figure~\ref{fig:annotation_prompt} shows the system prompt provided to each AI annotator alongside the episode and rule-based draft annotation. Each prompt includes task-specific context tailored to the respective environment; the Jericho version is shown as a representative example.

\begin{figure}[h]
\centering
\begin{adjustbox}{max width=0.5\textwidth}
\begin{minipage}{\textwidth}
\hrule\vspace{4pt}
\begin{verbatim}
You are annotating failures made by an AI agent playing a Jericho
text-based adventure game.

Input.
You will be given a snapshot: the complete trajectory of one episode,
including every action the agent took and every game response it
received, along with the final score.

Your Task.
You will be given a rule-based draft annotation alongside the episode
snapshot. Your job is to produce a revised, complete annotation by:
(1) Reviewing the draft: correct any inaccurate type, where, or why
    fields; remove false positives.
(2) Completing missing fields: fill in tier (core/marginal) and
    location for every entry.
(3) Adding missed failures: identify any failures not caught by the
    rule-based classifier and add them to the list.

What Counts as a Failure.
A failure is any agent action (or sequence of actions) that causes
the agent to:
- Die (health reaches zero, or the game explicitly states the agent
  has died)
- Get stuck at a location with no way to progress
- Enter a repetitive loop with no meaningful progress

Defined Failure Modes.
Use the following taxonomy when labeling failures. Each failure must
be assigned a type from this list. If a failure does not fit any
existing type, you may define a new one (see Adding New Failure Modes
below).

{{ LIST OF FAILURE MODES }}

Note: perception_error and decision_error are mutually exclusive for
the same milestone -- report only one.

Adding New Failure Modes.
If a failure clearly does not fit any of the types above, you may
define a new one. Use the format <class>/<name>, where <class>
is one of system, strategy, or operation. Include a one-sentence
definition in the note field. New types will be reviewed and
potentially added to the taxonomy.

Core vs. Marginal.
Core failure (at most 3 per episode):
- Directly caused the agent to stop making meaningful progress
- Fixing it alone would likely unlock significant further exploration
- Occurs earlier in the episode (earlier blockages matter more, since
  later content becomes unreachable once blocked)

Marginal failure:
- Had minor or indirect impact on the outcome
- The agent recovered or continued meaningfully after it occurred
- Is a downstream consequence of a core failure, not an independent
  cause

When uncertain, prefer fewer core failures.

Output Format.
Return a JSON array. Each element represents one identified failure:
[
  {
    "failure_id": "short_snake_case_identifier",
    "type": "class/subtype",
    "tier": "core" | "marginal",
    "location": "Brief description of the game location and context",
    "why": "Root cause: what the agent did wrong and what it should
            have done instead",
    "where": [first_step, last_step],
    "note": "(optional) required if type is newly defined"
  }
]

- failure_id: short descriptive identifier, e.g. "lamp_not_lit"
- location: e.g. "West of House, before entering the cellar"
- why: explain both the mistake and the correct action
- where: 0-indexed step indices in the trajectory

If no failures are detected, return an empty array [].

Notes.
- Annotate all failures you can identify, not just the most obvious.
- The why hint is for analysis purposes -- it will not be shown to
  the agent being evaluated.
- Do not infer failures not evidenced in the trajectory.
\end{verbatim}
\vspace{2pt}\hrule
\end{minipage}
\end{adjustbox}
\caption{System prompt provided to AI annotators in Jericho games. List of failure modes are provided in Appendix~\ref{app:taxonomy}}.
\label{fig:annotation_prompt}
\end{figure}

\subsection{Inter-Annotator Agreement}
\label{app:agreement}

Table~\ref{tab:agreement} reports inter-annotator agreement between Claude Sonnet 4.6 and Gemini 2.5 Flash across all six tasks.
Agreement is restricted to failures both annotators independently labeled \textit{core}.
Two failures from different annotators are considered matched if their step ranges have IoU $\geq 0.5$; matching is performed greedily in descending order of IoU.
Localization is reported as F1 over matched pairs; failure mode as Cohen's $\kappa$; tier as Jaccard similarity of core failure instances.

\begin{table}[h]
\centering
\begin{adjustbox}{max width=\columnwidth}
\begin{tabular}{lcccccc}
\toprule
 & \textbf{Jer.} & \textbf{AW} & \textbf{BAI} & \textbf{SW} & \textbf{BWS} & \textbf{GTP} \\
\midrule
Localization (F1)       & 0.645 & 0.732 & 0.525 & 0.506 & 0.714 & 0.533 \\
Failure Mode ($\kappa$) & 0.297 & 0.403 & 0.230 & 0.525 & 0.641 & 0.433 \\
Tier (Jaccard)          & 0.643 & 0.637 & 0.726 & 0.664 & 0.674 & 0.853 \\
\bottomrule
\end{tabular}
\end{adjustbox}
\caption{Inter-annotator agreement (Claude Sonnet 4.6 vs.\ Gemini 2.5 Flash) across all six tasks. Jer.=Jericho, AW=AlfWorld, BAI=BabyAI, SW=ScienceWorld, BWS=Bundled Web Shopping, GTP=Group Travel Planning.}
\label{tab:agreement}
\end{table}

Agreement is generally high across tasks.
The exception is failure mode $\kappa$ for Jericho and BabyAI, which is noticeably lower than the other tasks.
Both tasks have more complex failure mode taxonomies, which makes it harder for two AI annotators to independently assign the same label.
This is addressed by the human expert adjudication step described in Section~\ref{sec:dataset}, which resolves conflicts in failure mode assignments for all core failure instances.

\section{Dataset}
\label{app:dataset}

\subsection{Failure Mode Taxonomy}
\label{app:taxonomy}

\begin{figure*}[!t]
  \centering
  \includegraphics[width=\textwidth]{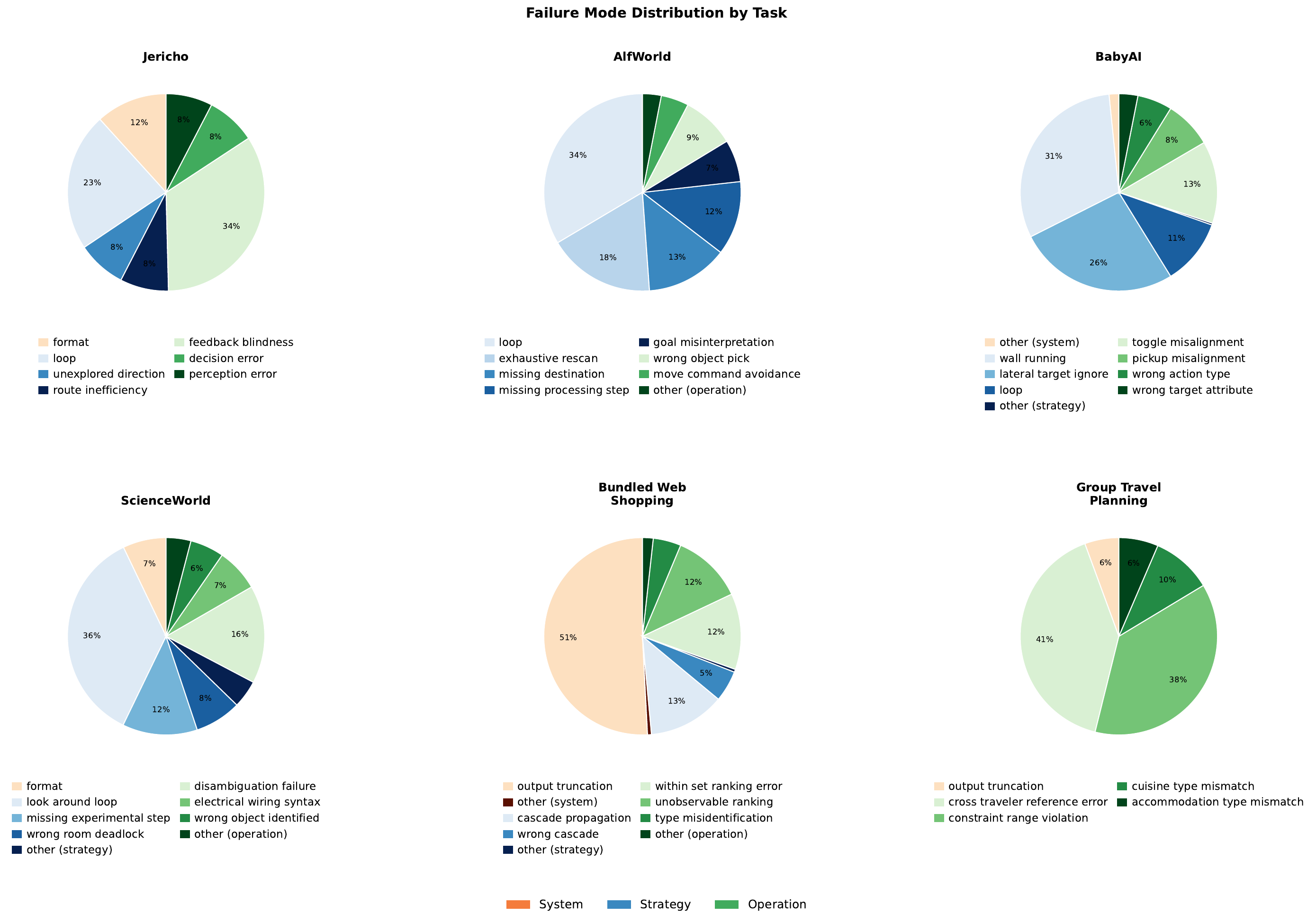}
  \caption{Failure mode distribution across the six BenchTrace tasks. Colors indicate failure class: system (orange), strategy (blue), and operation (green).}
  \label{fig:failure_mode_dist}
\end{figure*}

Table~\ref{tab:failure_modes} shows the failure mode taxonomy for Jericho as a representative example; taxonomies for all other tasks are provided in the project repository.
Figure~\ref{fig:failure_mode_dist} shows the failure mode distribution across all six tasks.

\begin{table*}[h]
\centering
\begin{adjustbox}{max width=\textwidth}
\begin{tabular}{lp{6cm}p{5cm}}
\toprule
\textbf{Type} & \textbf{Definition} & \textbf{Localization} \\
\midrule
\multicolumn{3}{l}{\textit{System}} \\
\midrule
\texttt{system/format} & Agent outputs a malformed action (empty, multi-line, JSON leakage, or excessively long) & Step where the malformed action occurs \\
\midrule
\multicolumn{3}{l}{\textit{Strategy}} \\
\midrule
\texttt{strategy/loop} & Within a no-progress segment, the agent repeats the same (room, action) pair & First repeated (room, action) pair; range covers the full loop \\
\texttt{strategy/route\_inefficiency} & Steps between two consecutive milestones exceed 2$\times$ the walkthrough reference & From the step reaching milestone $N{-}1$ to the step reaching milestone $N$ \\
\texttt{strategy/unexplored\_direction} & An available exit is mentioned in observations but never tried across multiple visits & From the first step the agent passes through the room without trying the direction to the end of the episode \\
\midrule
\multicolumn{3}{l}{\textit{Operation}} \\
\midrule
\texttt{operation/feedback\_blindness} & Agent repeats the same action after receiving negative feedback for it & Step of the first repeated occurrence after negative feedback \\
\texttt{operation/perception\_error} & Agent enters the correct room for a milestone but never attempts the correct action & First step where the agent enters the correct room \\
\texttt{operation/decision\_error} & Agent attempts the correct action at the wrong time, or uses the correct verb with the wrong object & Step where the incorrect attempt occurs \\
\bottomrule
\end{tabular}
\end{adjustbox}
\caption{Failure mode taxonomy for Jericho (representative example).}
\label{tab:failure_modes}
\end{table*}

\subsection{Dataset Entry Format}
\label{app:dataset_format}

Each dataset entry is a JSON object with the fields described in Table~\ref{tab:dataset_format}.

\begin{table}[h]
\centering
\begin{adjustbox}{max width=\columnwidth}
\begin{tabular}{llp{4.5cm}}
\toprule
\textbf{Field} & \textbf{Type} & \textbf{Description} \\
\midrule
\multicolumn{3}{l}{\textit{Top level}} \\
\midrule
\texttt{id} & string & Unique episode identifier. \\
\texttt{game} & string & Task name, e.g.\ \texttt{detective} \\
\texttt{model} & string & Model that produced the episode, e.g.\ \texttt{gpt-4.1} \\
\texttt{snapshot} & object & Full episode state (see below) \\
\texttt{failure\_instances} & object & Annotated failure instances (see below) \\
\midrule
\multicolumn{3}{l}{\textit{snapshot}} \\
\midrule
\texttt{final\_score} & float & Score achieved at episode end \\
\texttt{max\_score} & int & Maximum achievable score for the task \\
\texttt{n\_steps} & int & Total number of steps in the episode \\
\texttt{trajectory} & list & Sequence of steps; each step contains \texttt{step} (0-indexed), the information received from the environment, and the action taken by the agent \\
\midrule
\multicolumn{3}{l}{\textit{failure\_instances}} \\
\midrule
\texttt{core\_failure} & list & Human-adjudicated core failures. Each entry contains: \texttt{type} (failure mode from taxonomy), \texttt{where} (step range \texttt{[first, last]}, inclusive), and \texttt{diagnosis} (one-sentence description of the root cause and correct action) \\
\texttt{marginal\_failure} & list & AI-identified marginal failures not confirmed as core. Each entry contains \texttt{type} and \texttt{where} only \\
\bottomrule
\end{tabular}
\end{adjustbox}
\caption{Fields of each dataset entry.}
\label{tab:dataset_format}
\end{table}


\section{Evaluation Suite}
\label{app:experiments}

\subsection{Prompt for Reflection Evaluation}
\label{app:reflection_prompts}

Figure~\ref{fig:reflection_prompts} shows the reflection evaluation prompts for Jericho. Prompts for other tasks are adapted to their respective observation formats and are available in the project repository. All three questions share the same system prompt, which instructs the model to evaluate an agent playing a Jericho text adventure game. Each question prompt is additionally filled with the episode and metadata at inference time.

\begin{figure}[h]
\centering
\begin{adjustbox}{scale=0.5}
\begin{minipage}{\textwidth}
\hrule\vspace{4pt}
\textbf{System prompt (shared across detection--diagnosis)}
\begin{verbatim}
You are evaluating a language model agent playing a Jericho text adventure game.

In each step of the episode, the agent receives:
- obs: the game's text output (description of the current situation)
- inv: the agent's current inventory
- action: the command the agent issued in response

The episode ends after 110 steps. The agent earns points by completing objectives.

You will be shown the full trajectory (all 110 steps) along with the agent's
final score and the maximum achievable score. Your task is to analyze the
agent's performance and answer specific questions about what went wrong and where.
\end{verbatim}
\vspace{4pt}\hrule\vspace{8pt}
\textbf{detection}
\begin{verbatim}
Here is the trajectory of an agent playing {game}. The agent achieved a final
score of {final_score} / {max_score} points.

<trajectory>
{trajectory}
</trajectory>

Does this agent's performance have room for improvement?

Respond with a JSON object only, no other text:
{"answer": "yes"} or {"answer": "no"}
\end{verbatim}
\vspace{4pt}\hrule\vspace{8pt}
\textbf{localization}
\begin{verbatim}
Here is the trajectory of an agent playing {game}. The agent achieved a final
score of {final_score} / {max_score} points.

<trajectory>
{trajectory}
</trajectory>

Identify up to 3 step ranges where improving the agent's behavior would most
effectively boost its performance. Focus on the most impactful issues.

Respond with a JSON array only, no other text. Each element must have
"step_start" and "step_end" (inclusive step numbers). Example:
[{"step_start": 10, "step_end": 25}, {"step_start": 60, "step_end": 70}]
\end{verbatim}
\vspace{4pt}\hrule\vspace{8pt}
\textbf{diagnosis}
\begin{verbatim}
Here is the trajectory of an agent playing {game}. The agent achieved a final
score of {final_score} / {max_score} points.

<trajectory>
{trajectory}
</trajectory>

The agent performed poorly in steps {step_start}--{step_end}. In one sentence,
describe what went wrong in this specific step range.

Also classify the failure using exactly one of these types:
- system/format: agent issued syntactically malformed or unrecognized commands
- strategy/loop: agent repeated the same (location, action) pairs without progress
- strategy/route_inefficiency: agent took an unnecessarily long or redundant path
- strategy/unexplored_direction: agent failed to explore an available direction
- operation/feedback_blindness: agent ignored explicit negative feedback
- operation/perception_error: agent failed to notice or interpret a key object
- operation/decision_error: agent made a logically incorrect decision
- unknown: the failure does not clearly fit any of the above

Respond with a JSON object only, no other text:
{"failure_type": "...", "description": "..."}
\end{verbatim}
\vspace{2pt}\hrule
\end{minipage}
\end{adjustbox}
\caption{Prompts used in the Reflection Evaluation (detection--diagnosis). Placeholders in braces are filled with episode-specific values at inference time.}
\label{fig:reflection_prompts}
\end{figure}

\subsection{LLM-as-Judge Prompt for Diagnosis}
\label{app:judge_prompt}

Figure~\ref{fig:judge_prompt} shows the LLM-as-judge prompt for diagnosis for Jericho. Prompts for other tasks are adapted to their respective observation formats and are available in the project repository. The judge is given the ground-truth diagnosis and the model's predicted diagnosis for the same step range, and assigns a raw score of 0, 1, or 2. The reported LLM-Judge score is this raw score divided by 2, normalized to $[0, 1]$.

\begin{figure}[h]
\centering
\begin{adjustbox}{scale=0.5}
\begin{minipage}{\textwidth}
\hrule\vspace{4pt}
\textbf{System prompt}
\begin{verbatim}
You are evaluating the quality of a failure diagnosis produced by a language model.
You will be given a ground-truth diagnosis and a predicted diagnosis for the same
step range of an agent episode. Your task is to judge whether they describe the
same underlying failure.
\end{verbatim}
\vspace{4pt}\hrule\vspace{8pt}
\textbf{User prompt}
\begin{verbatim}
Ground-truth diagnosis: {ground_truth}

Predicted diagnosis: {prediction}

Both diagnoses concern steps {step_start}--{step_end} of the same episode.

Score the predicted diagnosis on a scale of 0--2:
  2 = Correct: describes the same root cause and the same agent mistake as the
      ground truth (minor wording differences are acceptable).
  1 = Partial: partially overlaps with the ground truth --- identifies the correct
      general area of failure but misses the specific root cause or correct action.
  0 = Incorrect: describes a different failure, or is vague to the point of being
      uninformative.

Respond with a JSON object only, no other text:
{"score": <0|1|2>, "reason": "one sentence justifying the score"}
\end{verbatim}
\vspace{2pt}\hrule
\end{minipage}
\end{adjustbox}
\caption{Prompt used for LLM-as-Judge scoring in Diagnosis. Placeholders in braces are filled at inference time.}
\label{fig:judge_prompt}
\end{figure}

\subsection{Non-Generalizing Reflection: A Case Study}
\label{app:failure_patterns}

We illustrate how an agent's reflection may not generalize beyond the snapshot in which it was produced, using two reflections on a subtask from Jericho (Figure~\ref{fig:generalize_example}).
The non-generalizable reflection encodes specific room names, directions, and death conditions from a particular trajectory, information that does not transfer when the snapshot changes.
The generalizable reflection instead extracts transferable principles applicable across different contexts.

\begin{figure}[h]
\centering
\begin{adjustbox}{scale=0.5}
\begin{minipage}{\textwidth}
\hrule\vspace{4pt}
\textbf{Non-generalizable reflection}
\begin{verbatim}
Mandatory Actions for Progress:
- Read the Chief's initial paper first to understand the murder timeline.
- From the Chief's office, always go north to the outside
  (death occurs if you enter the restaurant directly).
- From the outside, go west to the street (avoid east as it's a dead end).
- On the street, investigate the mayor's home (east) first.
- The restaurant (north) is a trap leading to death.

Avoid Death-Inducing Actions:
- Never enter the restaurant before checking the mayor's home.
- Do not flash your badge in hostile situations.

Current Game State Summary:
- If in the street, prioritize mayor's home.
- If in the mayor's home, search for murder evidence.
- If in the restaurant, reverse to the street immediately (death trap).
\end{verbatim}
\vspace{4pt}\hrule\vspace{8pt}
\textbf{Generalizable reflection}
\begin{verbatim}
1. When given an object by an NPC, immediately EXAMINE it to learn more.
2. After examining, GET the object if possible, to add it to your inventory.
3. READ any documents or papers you acquire for essential clues.
4. Follow explicit instructions from characters or the environment.
5. Upon entering a new area, use LOOK to review surroundings and exits.
6. Choose exits carefully: avoid areas that previously resulted in death.
7. Do not repeat unproductive actions (e.g., repeatedly LOOKing with no gain).
8. Prioritize exploring locations that appear safer or significant.
9. If unsure, re-examine rooms, check inventory for unread items,
   and plan to explore unvisited locations.
\end{verbatim}
\vspace{4pt}\hrule
\end{minipage}
\end{adjustbox}
\caption{Two reflections on the same subtask in \textit{Detective} (Jericho). The non-generalizable reflection memorizes snapshot-specific state; the generalizable reflection extracts transferable principles.}
\label{fig:generalize_example}
\end{figure}

\subsection{Failure Avoidance Rate (FAR) Computation}
\label{app:avoidance_rate}

We use a rule-based method to determine whether the agent avoids $f_d$.
Given a test snapshot truncated at truncation step $k$, the agent produces a post-truncation steps by continuing from step $k$ onward.
Avoidance is then determined by comparing this trajectory against $f_d$, with the criterion depending on the failure class.

\paragraph{System \& operation failures}
The agent avoids $f_d$ if none of its post-truncation steps produces the same environmental observation as any step within $f_d$'s step range, i.e., the agent does not perform the same erroneous operation in the same environment.

\paragraph{Strategy failures}
The agent avoids $f_d$ if the maximum observation recall over a sliding window against $f_d$'s step range remains below 0.5, where the sliding window length equals the length of $f_d$'s step range.

\paragraph{Human and LLM validation}
To verify the reliability of the rule-based computation, we sampled 30 test cases per task (20 labeled not avoided, 10 labeled avoided) across six tasks.
Claude Sonnet 4.6 and a human annotator independently judged avoidance for each sample; cases where both agreed (132 of 180) serve as ground truth.
The rule-based FAR labels achieve an accuracy of 81.8\% against this ground truth, with Cohen's $\kappa = 0.538$ (moderate agreement).
Agreement is higher for operation failures (88.6\%) than for strategy failures (74.2\%), consistent with the greater complexity of strategy failures.
These results confirm that the rule-based computation reliably tracks human judgment and scales to large evaluation sets without manual annotation.

\subsection{Task Difficulty}
\label{app:difficulty}

Table~\ref{tab:difficulty} reports agent progress and win rates across the six tasks, computed on snapshots before failure detection.
All post-detection snapshots contain at least one annotated failure instance, so applying failure detection would trivially yield win rates near zero and is therefore not informative.
Difficulty labels (Easy/Medium/Hard) are assigned based on these pre-detection rates and reflect how challenging each task is for a typical agent.

\begin{table}[h]
\centering
\begin{adjustbox}{max width=\columnwidth}
\begin{tabular}{lrrrrrr}
\toprule
 & \textbf{Jer.} & \textbf{AW} & \textbf{BAI} & \textbf{SW} & \textbf{BWS} & \textbf{GTP} \\
\midrule
Progress (\%)  & 23.5 & 52.5 & 50.6 & 29.7 & 24.2 & 94.9 \\
Win rate (\%)  &  0.0 & 17.1 & 13.9 &  8.4 &  0.3 & 41.8 \\
Difficulty     & \hard & \med & \med & \hard & \hard & \easy \\
\bottomrule
\end{tabular}
\end{adjustbox}
\caption{Agent progress and win rates before failure detection, and assigned difficulty labels. Jer.=Jericho, AW=AlfWorld, BAI=BabyAI, SW=ScienceWorld, BWS=Bundled Web Shopping, GTP=Group Travel Planning.}
\label{tab:difficulty}
\end{table}

\section{Experiments and Results}
\label{app:exp_results}

\subsection{Failure Repeat Count (FRC)}
\label{app:repeat_failure}

Table~\ref{tab:repeat_failure} reports the FRC across agent frameworks and backbone models.
A lower count indicates that the agent more effectively avoids reproducing the same failure after the truncation step.

\begin{table}[h]
  \centering
  \begin{adjustbox}{width=\columnwidth}
  \begin{tabular}{llcccc}
    \toprule
    \multirow{2}{*}{\textbf{Category}} & \multirow{2}{*}{\textbf{Framework}} & \multicolumn{2}{c}{\textbf{Jericho}} & \multicolumn{2}{c}{\textbf{AlfWorld}} \\
    \cmidrule(lr){3-4} \cmidrule(lr){5-6}
    & & \textbf{Q} & \textbf{G} & \textbf{Q} & \textbf{G} \\
    \midrule
    Non-evolution & ReAct     & 3.31 & 1.74 & 0.65 & 0.28 \\
    \midrule
    \multirow{6}{*}{Self-evolution}
      & RAG       & 4.41 & 1.80 & 0.57 & 0.44 \\
      & ReMem     & 3.25 & 1.96 & 0.63 & 0.41 \\
      & MemRL     & 4.72 & 1.27 & 0.59 & 0.43 \\
      & Reflexion & 1.89 & 1.64 & 0.59 & 0.46 \\
      & EvoTest   & 1.91 & 2.32 & 0.50 & 0.43 \\
      & AutoSkill & 1.99 & 1.32 & 0.57 & 0.30 \\
    \bottomrule
  \end{tabular}
  \end{adjustbox}
  \caption{FRC (failure repeat count) measures the average number of times an agent reproduces the target failure $f_d$ in the post-truncation steps ($\downarrow$). Q = Qwen3-32B; G = GPT-4.1.}
  \label{tab:repeat_failure}
\end{table}

\subsection{FAR by Failure Proximity and Evolution Distance on Jericho}
\label{app:type_distance}

Table~\ref{tab:type_distance} reports score\,/\,FAR for each combination of failure proximity and evolution distance on Jericho (Qwen3-32B).

\begin{table*}[t]
  \centering
  \begin{adjustbox}{width=\textwidth}
  \begin{tabular}{llccccccccc}
    \toprule
    & & \multicolumn{3}{c}{\textbf{Snapshot-level}} & \multicolumn{3}{c}{\textbf{Subtask-level}} & \multicolumn{3}{c}{\textbf{Task-level}} \\
    \cmidrule(lr){3-5}\cmidrule(lr){6-8}\cmidrule(lr){9-11}
    \textbf{Category} & \textbf{Framework} & $d=1$ & $d=5$ & $d=10$ & $d=1$ & $d=5$ & $d=10$ & $d=1$ & $d=5$ & $d=10$ \\
    \midrule
    Non-evolution & ReAct     & 23.2\,/\,23.3 & 28.0\,/\,32.1 & \textbf{27.0}\,/\,8.3 & 21.2\,/\,44.4 & 20.2\,/\,30.4 & 19.2\,/\,12.5 & \textbf{23.9}\,/\,26.7 & \underline{24.4}\,/\,23.3 & 18.7\,/\,20.0 \\
    \midrule
    \multirow{6}{*}{Self-evolution}
      & RAG       & 24.0\,/\,30.8 & \underline{28.4}\,/\,22.2 & 25.7\,/\,9.1 & \textbf{25.6}\,/\,36.0 & \textbf{25.9}\,/\,30.0 & 23.2\,/\,\textbf{22.2} & \underline{23.0}\,/\,25.9 & 24.0\,/\,29.0 & 18.8\,/\,20.0 \\
      & ReMem     & 25.8\,/\,\underline{37.0} & 27.0\,/\,35.7 & 23.4\,/\,\textbf{36.4} & 23.3\,/\,\textbf{57.1} & 22.9\,/\,\underline{36.4} & 22.2\,/\,12.5 & 21.8\,/\,\underline{40.7} & 22.8\,/\,\textbf{51.7} & 19.6\,/\,20.0 \\
      & MemRL     & 25.8\,/\,32.1 & 26.6\,/\,22.2 & 25.9\,/\,25.0 & \underline{25.5}\,/\,33.3 & \underline{23.1}\,/\,\textbf{39.1} & \underline{24.9}\,/\,0.0 & 21.8\,/\,37.0 & \textbf{24.6}\,/\,26.7 & 18.4\,/\,\textbf{30.0} \\
      & Reflexion & \underline{26.5}\,/\,26.7 & 27.6\,/\,\underline{37.5} & \underline{26.4}\,/\,16.7 & 21.6\,/\,33.3 & 21.2\,/\,\underline{36.4} & 22.3\,/\,12.5 & 20.1\,/\,28.6 & 21.6\,/\,29.0 & \textbf{20.8}\,/\,18.2 \\
      & EvoTest   & 23.7\,/\,\textbf{48.1} & 24.9\,/\,\textbf{40.9} & 22.4\,/\,16.7 & 24.3\,/\,\underline{50.0} & 22.4\,/\,27.3 & \textbf{31.3}\,/\,\textbf{22.2} & 21.0\,/\,30.0 & 23.2\,/\,\underline{32.3} & 18.4\,/\,9.1 \\
      & AutoSkill & \textbf{27.0}\,/\,25.9 & \textbf{28.7}\,/\,37.0 & 21.6\,/\,\underline{27.3} & 23.4\,/\,29.2 & 21.2\,/\,20.0 & 20.8\,/\,12.5 & 17.6\,/\,\textbf{46.2} & 24.2\,/\,26.7 & \underline{20.3}\,/\,\underline{25.0} \\
    \bottomrule
  \end{tabular}
  \end{adjustbox}
  \caption{Score\,/\,FAR (\%) on Jericho (Qwen3-32B) broken down by failure proximity and evolution distance~$d$.}
  \label{tab:type_distance}
\end{table*}


\end{document}